\PassOptionsToPackage{table}{xcolor}
\documentclass{article} 
\usepackage{iclr2021_conference,times}

\usepackage{amsmath,amsfonts,bm}

\def\eqref#1{equation~\ref{#1}}

\def\1{\bm{1}}

\DeclareMathAlphabet{\mathsfit}{\encodingdefault}{\sfdefault}{m}{sl}
\SetMathAlphabet{\mathsfit}{bold}{\encodingdefault}{\sfdefault}{bx}{n}

\usepackage{hyperref}
\usepackage{url}
\usepackage{tabularx} 
\usepackage[table]{xcolor} 
\usepackage{hhline} 
\usepackage{graphicx}
\definecolor{darkgreen}{rgb}{0.0, 0.5, 0.0}

\title{Evaluating Hallucinations in Domain-Adapted Large Language Models}

\author{Sanchita Porwal, Sai Prasath S, Xingjian Bi, \& Madelyn Scandlen \\
College of Computing\\
Georgia Institute of Technology\\
Atlanta, GA 30318, USA \\
\texttt{\{sporwal9,ss651,xbi38,mscandlen3\}@gatech.edu}
}
\iclrfinalcopy 
\begin{document}

\maketitle

\begin{abstract}
This study investigates the phenomenon of hallucinations in domain-adapted Large Language Models (LLMs), focusing on the fine-tuning of the Llama-2 model with the Lamini dataset. Hallucinations, or the generation of nonsensical or unfaithful content by LLMs, pose a significant challenge, especially when these models are fine-tuned with domain-specific data. Our methodology involves a series of experiments testing memorization, recall, and reasoning capabilities of the fine-tuned LLM, comparing its performance on novel question-answer pairs and domain-specific information. We found that while the model shows proficiency in tasks similar to its training data, its capability to accurately reason about and recall new domain-specific information remains limited, leading to instances of hallucination. The model demonstrates a tendency to provide correct answers with extra information, suggesting an inclination toward over-generation. These results suggest important limitations of fine-tuning-only approaches for mitigating hallucinations when adapting LLMs to specialized domains and underscores the need for more robust methods in adapting LLMs to specialized domains. The study also provides insights into the varying performance of LLMs on different types of information, revealing a comparative weakness in handling domain-specific queries.

\end{abstract}

\section{Introduction}
\label{intro}

Large Language Models (LLMs) such as ChatGPT \citep{chatgpt}, Bard (Google, 2023), and Llama-2 \citep{llama2} have become ubiquitous tools for a multitude of tasks across various domains. Tasks like summarization and question-answering on cloze-prompts have shown high performance, where language models’ facilitation of natural language can deceive humans into believing the outputs were also from humans \citep{Gudibande2023} \citep{bai2022}. With the boom of artificial intelligence, business organizations have utilized open-source LLMs through fine-tuning on organization-specific information as a tool for internal and external use \citep{arefeen2023leancontext}.

With growing popularity of LLMs, new challenges arise. The capabilities of language models have impressed the public but this can lead to a false sense of trust and reliance, which can be dangerous and vulnerable to adversarial uses. One common challenge of large language models is the tendency to hallucinate, defined as generated content that is nonsensical or unfaithful to the provided source content \citep{ji2023survey}. Models are trained to avoid hallucination by responding that the model does not have enough information to answer the question, but when hallucinations occur, there can be negative implications \citep{zhang2023sirens}.

Hallucinations are likely to occur when LLMs ingest new data \citep{mckenna2023sources}. LLMs may lack the knowledge to answer questions about all information due to the restriction of training. Retrieval-augmented generation (RAG) and fine-tuning have been proposed as solutions to introduce new information to LLMs \citep{brown2020fewshot}. However, research has shown that fine-tuning LLMs may not address the issue of hallucinations \citep{Gudibande2023}. 

In this work, we evaluate the robustness of LLMs against hallucination on new information through a series of recall, memorization, and reasoning experiments. We fine-tune Llama-2 on question-answer pairs containing new information unseen during pre-training. 

To test memorization, we first evaluated the fine-tuned LLM’s ability to correctly answer questions of the test dataset, finding that it answered the questions most similar to those in the training set, seen during fine-tuning. To address recall, we designed multi-hop question-answer pairs using Chat-GPT and evaluated the responses from the fine-tuned model at synthesizing information across a more complex question. For reasoning, we performed a series of experiments inspired to test the specific capabilities of understanding, retrieval, stress test and reasoning on domain-specific information and general information. We address the question: \textbf{``do LLMs actually learn new domain-specific information during fine-tuning or does it lead to hallucinations?"} 

When compared with common-knowledge information likely seen during pre-training, our results suggest that the fine-tuned model does not reliably reason about newly introduced domain-specific material, making it susceptible to hallucination in this setting.

\section{Related Work}
\label{related}

\subsection{Domain Specific Fine-tuning}
Large Language Models (LLMs) typically exhibit strong performance in a broad array of tasks and areas, but their effectiveness in specific domains is often constrained by the scarcity of specialized training data. Domain specific fine-tuning has shown to improve the performance of language models \citep{gururangan2020don}. \citet{zheng2023trafficsafetygpt} fine-tuned the Llama2-7B model for creating a transportation safety expert which outperforms ChatGPT in terms of BLEU and ROUGE scores. Similarly, \citet{liu2023chipnemo} built a LLM for chip design by fine-tuning Llama2-7B and 13B models which outperforms the corresponding models by around 20\% in terms of accuracy of chip design tasks. 

\subsection{Introducing New Information}
While domain specific fine-tuning of LLMs has shown significant improvements in performance, introducing new knowledge during the fine-tuning phase can lead to the model hallucinating on the newly introduced entities. \cite{Schulman_2023} showed that introducing new knowledge during fine-tuning is equivalent to adding new nodes to the pre-trained knowledge graphs of an LLM, and can lead to creation of weak links (affects reasoning) and limited comprehension (affects understanding). Moreover, \cite{Gudibande2023} showed that introducing knowledge during fine-tuning can lead to the model guess or hallucinate answers, rather than actually doing the task. However, none of these prior studies quantitatively evaluate hallucinations in LLMs and more specifically the ability of LLMs to learn newly introduced knowledge during fine-tuning. 


\subsection{Hallucination}

\cite{ji2023survey} defined hallucination as “generated content that is nonsensical or unfaithful to the provided source content”. They also classified hallucinations into two categories: (1) intrinsic: where the generated output directly contradicts the source content, and (2) extrinsic: where the generated output neither be supported not be contradicted by the source context, and external information is required to validate the output. \cite{zheng2023does} studied the hallucinations in ChatGPT and classified them into three categories: (1) memorization: where the model does not learn the information present in the training dataset, (2) recall: where the model learns but can't recall the required information, and (3) reasoning: where the model can recall but can't reason between the recalled entities. For this study, we use the nomenclature proposed by \cite{zheng2023does} for categorizing hallucinations. \\

Since the large scale adoption of LLMs, much work has been done to mitigate the threat of hallucinations in generated responses. \citet{dhuliawala2023chainofverification} proposed Chain-of-Verification (COVE) method where the model generates an initial response, then formulates and answers its own validation questions to verify the accuracy of the initial response. \cite{Lewis2021} proposed Retrieval Augmented Generation (RAG) where the LLM is augmented with relevant context retrieved from a set of documents. These techniques have shown to improve model performance and reduce hallucinations in LLMs. While this study focuses on evaluating hallucinations and not mitigating them, we still use RAG for generating gold labels for few LLM tasks.




\section{Dataset and Model} 
\subsection{Dataset description}
In this paper, we fine-tune using the Lamini-docs dataset \citet{lamini}. It contains a comprehensive collection of 1,260 question-answer pairs curated to encompass a wide range of documentation about the application of APIs, functions, variables, and parameters within the context of the Lamini Library. This dataset includes 140 examples in the test subset.

The rationale behind the selection of the Lamini dataset is that (1) it is relatively new and (2) introduces new function names and API calls. There is a higher probability that the data constituting the Lamini dataset were not encountered by the model during its pre-training phase. Consequently, this helps a more accurate assessment of the model's propensity for generating hallucinations, particularly when confronted with newly introduced information that it has not previously encountered. 

\subsection{Model description}

We use the ``Llama-2-7b-chat-hf" \citep{llama2} as our base model because of its pre-existing chat functionalities. Moreover, given the comparatively small scale of our training dataset, it is more efficient to fine-tune an already established chat model rather than fine-tuning a pre-trained model. \\
\\
We fine-tune the base model using the QLoRA technique \citep{dettmers2023qlora} with 4 bit quantization. We fine-tune all the linear layers with a LoRA rank of 64, fine-tuning only 1.16\% of the total model parameters. We fine-tune the model over 4 epochs with a learning rate of 5e-5 and a batch size of 16. Detailed hyper-parameters employed are detailed in Table~\ref{hyper}.

\newcolumntype{C}[1]{>{\centering\arraybackslash}p{#1}}
\begin{table}[!htb]
  \centering
  \caption{QLoRA fine-tuning Hyper-parameters}
  \vspace{3mm}
  \begin{tabular}{|C{4cm}|C{2cm}|}
    \hline
    \textbf{Hyperparameter} & \textbf{Value} \\
    \hline
    Quantization  & 4 bit \\
    \hline
    Rank & 64 \\
    \hline
    Alpha & 16 \\
    \hline
    Epochs & 4 \\
    \hline
    Learning rate & 5e-5 \\ 
    \hline
    Batch Size & 16 \\
    \hline
  \end{tabular}
  \vspace{10pt} 
  \label{hyper}
\end{table}

\section{Methodology}
\label{method}

\subsection{Annotating Model Responses}
\label{sec:annotation}
For evaluating the hallucinations, we compare the fine-tuned model generated responses with the gold responses and categorize them into three categories:
\begin{enumerate}
    \item \textbf{Wrong:} The generated model answer does not align or only partially aligns with the gold responses. 
    \item \textbf{Correct + Extra Details: } The generated model answers aligns with the gold response, but includes additional relevant information which is not necessary. 
    \item \textbf{Correct:} The generated model answer aligns with the gold response without including any extra details. 
\end{enumerate}

We use GPT-4 to perform the annotations by using the following prompt: \textit{``Given the gold answer and the generated answer, categorize them into: \textless{}annotation description\textgreater{} \\Gold answer: \textless{}gold answer\textgreater{}\\Generated answer: \textless{}generated answer\textgreater{} "}

\subsection{Memorization}
\label{sec:mem}
For evaluating the model's memorization abilities, we focused on analyzing the fine-tuned model's performance on the test set by prompting it with questions and comparing the generated answers against the expected gold responses. Due to inconsistencies between the train set and test set, the study involved a double comparison method: each generated answer was compared not only to the actual test answer but also to the most similar train question's answer. The most similar train question was determined using the highest Sentence-BERT \citep{reimers2019sentence} similarity between the train and test questions. This approach aimed to enhance the reliability of the evaluation process.

We evaluate the correctness of the responses using the annotation system discussed in Section \ref{sec:annotation}. Other than correctness, we also analyze other properties of the generated response: (1) length of the response, and (2) frequency of domain-specific terms in the answer. The technique for collecting the domain-specific terms is discussed in Section \ref{sec:new}. 



\subsection{Recall}
\label{sec:recall}
For evaluating the recall capabilities of the fine-tuned model,  we create a set of complex questions that mimic the style of multi-hop question-answering (MHQA) \citep{tang-etal-2021-multi}. MHQA necessitates that a model retrieves and synthesizes information across multiple textual passages to formulate a coherent response to a query.

We manually selected a subset of question-answer pairs from our test set. These pairs were then input to GPT-4 \citep{gpt4} for generating MHQA style questions and answers. We refer to the initial QA pairs as $Q_1 - A_1$ and $Q_2-A_2$. We then prompt GPT-4 using the prompt for generating $Q_3 - A_3$: \textit{``I'll give you two question-answer pairs, use the information in both q-a pairs and generate a new question and a new answer. The newly generated question should have 15-30 words. The generated answer should have less than 80 words"}. The newly created question \( Q_3 \) is designed to reflect the content of both \( Q_1 \) and \( Q_2 \), and the answer \( A_3 \) merges the information from \( A_1 \) and \( A_2 \). Thus, The fine-tuned model must demonstrate a comprehensive understanding of the original questions to accurately address \( Q_3 \).




To evaluate the recall ability of the fine-tuned model, we prompt the model with $Q_3$ and compare the generated response with the gold response $A_3$. Similar to the evaluation in Section \ref{sec:mem}, we analyze the correctness of the responses using the annotation system proposed in Section \ref{sec:annotation}, along with other properties of the responses: (1) length, and (2) frequency of domain-specific entities.

\subsection{Reasoning}

During domain specific fine-tuning, new entities/terms are often introduced as part of the fine-tuning dataset which weren't present in the initial pre-traning dataset. While the high level analysis helps with evaluating the overall ability of the fine-tuned LLM in answering questions, the low level analysis helps evaluate the ability of LLMs to actually learn these newly introduced entities (or terms) by performing a more fine-grained analysis. 
\subsubsection{Filtering Newly Introduced Entities}
\label{sec:new}
For performing the low level analysis, we first extract the newly introduced entities from the fine-tuning dataset using a three stage filtering process:
\begin{enumerate}
    \item \textbf{TF-IDF Words:} We collect the top 1000 tf-idf words along with their ranks (based on tf-idf score) after removing the English stop words. This filtering stage helps select the important words along with their ranks, and discard stop words and other frequently occurring words. 
    \item \textbf{Remove GloVe Words:} In the second filtering stage, we remove all the GloVe words present in the top-1000 tf-idf words. We use the GloVe 6B embeddings generated using the Wikipedia 2014 and Gigaword 5 dataset. Removing words present in this corpus ensures that we select only the domain specific entities and remove the common words.
    
    \item \textbf{Manual Filtering:} To specifically analyze the model's learning of function calls, variable names, parameters, and Python scripts, we manually filter out these entities. Our focus is exclusively on these entities, disregarding other domain-specific entities, including those that may have been introduced during the fine-tuning process.
\end{enumerate}

The resultant entities from the three stage filtering process are then analyzed further. We interchangeably refer to these filtered entities as ``functions" or ``entities" in the rest of the paper. 

\subsubsection{Evaluating Reasoning}
\label{sec:eval_hal}
For studying whether these newly introduced terms (refer \ref{sec:new}) are actually learned by the model or whether the model hallucinates and guesses on these terms, we perform a four stage evaluation process:

\textbf{A. Understanding:} We evaluate whether the model understands the newly introduced entities by asking it to generate description for the functions. 
\begin{itemize}
    \item As the gold function descriptions are not available, we generate them using RAG and the base model. We collect all the QA pairs which contains the given function name, and provide it as context to base model using the prompt: \textit{``\textless{}context\textgreater{} Summarize the function \textless{}function\_name\textgreater{} in 30 words".} The generated function description are considered as the gold label as the base models are really good at summarizing \citep{zhang2023benchmarking}.
    \item We prompt the fine-tuned model with the question: \textit{``Generate a description for the \textless{}function\_name\textgreater{} function/parameter in 30 words."}.
    \item We then manually compare the function descriptions of the fine-tuned model and the RAG + base model, and annotate them using the scoring metric proposed in Section \ref{sec:annotation}.
\end{itemize}

\textbf{B. Retrieval: } Next, we evaluate the ability of the fine-tuned model to select the right function call for a given task. 

\begin{itemize}
    \item For creating the tasks, we use the gold function descriptions generated using RAG + base model, and prompt GPT-4 to create a question where the answer is the function call using the prompt: \textit{``\textless{}gold function description\textgreater{} Create a question using the above context where the answer is \textless{}function\_name\textgreater{}".} We refer to the generated questions as ``inverted questions". 
    \item The fine-tuned model is then prompted with these inverted questions for evaluating their zero shot retrieval abilities. 
    \item For each task, the fine-tuned model is assigned a score of 1 if the model correctly retrieves the function call, that is the model's output contains the function call. Else, the fine-tuned model is assigned a score of 0. 
\end{itemize}

\textbf{C. Stress Test: } To ensure that the model actually learns these newly introduced entities and not use the function names to deduce their descriptions, we perform a stress test where we ask the model to choose between multiple similar questions for a given task.

\begin{itemize}
    \item We reuse the inverted questions generated in ``B. Retrieval" for the stress test. However, instead asking the fine-tuned model to directly retrieve a function call for the task, we provide the fine-tuned model with multiple similar options and ask the model to choose the right one. We use GPT-4 for generating the similar options using the prompt: \textit{``\textless{}gold function description\textgreater{} What are 4 other function names the can be used instead of \textless{}function\_name\textgreater{}. Output as a python list."} We refer to the generated function names as ``similar function names". 
    \item We prompt the fine-tuned model using the prompt: \textit{``\textless{}inverted question\textgreater{} \textless{}similar function names\textgreater{} \textless{}actual function name\textgreater{}"}. Note that, we combined the similar function names with the actual function name, and shuffle the list so that the positions of the options are random. 
    \item We evaluate the model performance by analyzing whether the model selected the correct answer from the provided options. 
\end{itemize}

This test was inspired by the False Confidence Test and the None of the Above test proposed by \citet{umapathi2023med} which assess the LLMs ability to reason and provide logical and factual correct responses without creating fake information. \\

\textbf{D. Advanced Reasoning: }Finally, we evaluate the ability of the fine-tuned model to reason by studying whether it can recognize similarities between the newly introduced entities and the pre-training entities. 

\begin{itemize}
    \item For each of the new entities introduced, we carefully select equivalent entities from other platforms, including Google Cloud AI, Microsoft Azure AI, AWS AI service, IBM Watson Studio, and H2O.ai. Alongside these equivalent entities, we also document the rationale behind their equivalence. We refer to rationale as "justification". We ensure that the base model has knowledge of these selected equivalent entities by prompting them and evaluating their function descriptions. 
    \item To evaluate the fine-tuned model's reasoning abilities, we prompt the model with two questions:
    (1) \textit{``Are both \textless{}newly introduced entity\textgreater{} and \textless{} equivalent entity\textgreater{} used for \textless{}justification\textgreater{}?"} and (2) \textit{``Are there any similarities between \textless{}newly introduced entity\textgreater{} and \textless{} equivalent entity\textgreater{}?"} 
    \item We use two questions to ensure that the model has recall on both the entities, and only suffers from reasoning. (a) If the model answers the first question wrong, then the model does not have recall. (b) Whereas, if the model answers the first question right and gets the second question wrong the model has no reasoning. (c) If the model gets both the questions right, the model has both recall and reasoning. (d) The case where the model gets the first question wrong and the second question right can mean the model is either hallucinating or it has found a justification different from the one manually selected to link the entities. For evaluating the reasoning ability of the model, we only consider cases (b) and (c). 
\end{itemize}

Note that, we conduct the different stages of evaluation sequentially, meaning only those entities that successfully pass each preceding stage are considered for evaluation in the subsequent stage. We have ordered the stages based on their importance and relative difficulty levels. 





\section{Results}

\subsection{Memorization}
\label{sec:r-mem}
\begin{figure}[h]
    \centering
    \includegraphics[width=0.6\textwidth]{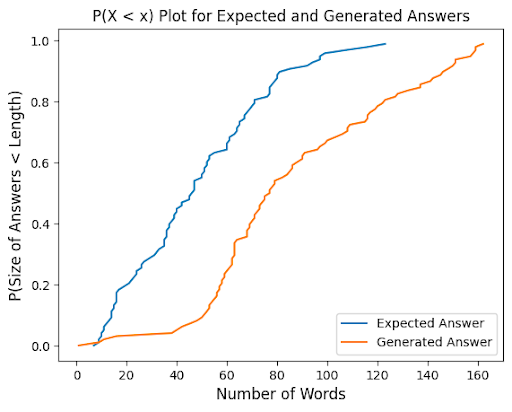}
    \caption{The fine-tuned model outputs longer answers than expected}
    \label{fig:image1}
\end{figure}

\textbf{Length:} Comparing the length of the generated responses with the expected response, we find that in 81.6\% of the examples the generated response is longer than the expected response. Moreover, generated responses (= 86.1 words per response) are almost twice as long as the expected responses (= 47.6 words per response). Figure \ref{fig:image1} shows the cdf of the expected and the generated responses with respect to the number of words in the response, we can observe that generated responses are much longer on average. Notably, none of the expected responses contain more than 123 words, however, 18.4\% of the generated responses are over 123 words. 

\begin{figure}[h]
    \centering
    \includegraphics[width=0.6\textwidth]{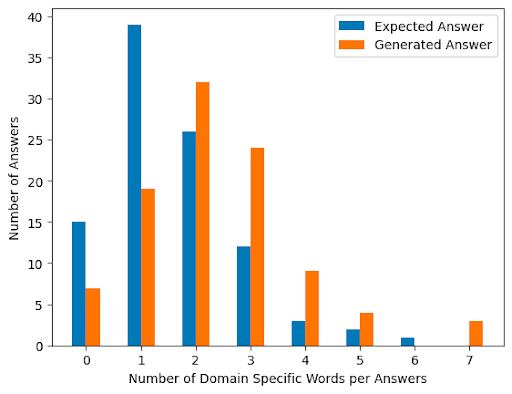}
    \caption{The fine-tuned model uses more domain specific words than expected}
    \label{fig:image2}
\end{figure}

\textbf{Use of domain specific words:} Comparing the use of domain specific words between the expected and the generated response, we find that generated responses contain 2.36 domain specific words per response, whereas, the expected answers contain 1.58 domain specific words per response. The top-5 domain specific words used by the fine-tuned model are: \{lamini, llms2, chatbots, llamaindex, hyperparameters\}. The corresponding ranks of these entities in the domain-specific entities are: 1, 2, 6, 8, and 10. Refer Section \ref{sec:new} for collecting domain specific words and their tf-idf ranks. 

\textbf{Correctness:} We use the annotation system discussed in Section \ref{sec:annotation} for comparing the expected and generated responses. We find that 43.88\% of the time, the model correctly answered the questions, and in 18.36\% of cases, the responses were wrong. Interestingly, 37.75\% of the correct answers included extra details not present in the expected answers. \\
\\
The average BERT similarity between the test question and the (most-similar) train question is 0.8856, which is very high. Therefore, we can observe that the test set questions are very similar to the train set.

An example of where the fine-tuned model generated additional content in addition to what was present in the actual answer:

 \textit{\textbf{Question:} Can I fine-tune the pre-trained models provided by Lamini on my own dataset? How does that process work? \\
 \\
 \textbf{Expected Answer:} \textcolor{darkgreen}{Yes, you can fine-tune the pre-trained models provided by Lamini on your own dataset \ldots} \\
 \\
    \textbf{Generated Answer:} \textcolor{darkgreen}{Yes, you can fine-tune the pre-trained models provided by Lamini on your own dataset \ldots} \textcolor{red}{It is important to note that fine-tuning a pre-trained model can be computationally intensive and may require significant resources, such as GPU power and large amounts of data. However, fine-tuning a pre-trained model can lead to significant improvements in performance and can be a cost \ldots}}

Hence, we find that fine-tuned domain specific models generate longer responses while memorizing and using important domain specific words. Moreover, the fine-tuned model correctly answers 80\% of the questions, while including extra details in half of those responses. 

\subsection{Recall}
\label{sec:r-recall}
\newcolumntype{C}[1]{>{\centering\arraybackslash}p{#1}}
\begin{table*}[!htb]
\centering
\renewcommand{\arraystretch}{1.7}
\caption{Examples of fine-tuned model generated answer}
\vspace{2mm}
\begin{tabular}{|p{5cm}|p{5cm}|p{5cm}|}
\hline
\textbf{GPT-4 Generated Question (Q\_3)} & \textbf{GPT Generated Answer (A\_3)} & \multicolumn{1}{|c|}{\textbf{Generated Answer}} \\ 
\hline

\textit{Can Lamini, which primarily operates with text data, effectively incorporate and process inputs from varied data types, such as audio and video, for content generation?} & \textit{\textcolor{darkgreen}{...Processing video content or generating outputs from video data might require additional tools or integration with specialized video processing software, as Lamini itself is not inherently designed for direct video processing tasks.}} & \textit{\textcolor{red}{Yes, Lamini can effectively incorporate and process inputs from varied data types, such as audio and video, for content generation. Lamini's LLM Engine can be trained on a variety of data types, including text, audio, and video, to generate high-quality content...}} \\
\hline
\textit{For troubleshooting and optimization of Lamini AI models, what resources are available to developers?}	&  \textit{\textcolor{darkgreen}{Developers have access to multiple resources for troubleshooting and optimizing Lamini AI models, including direct technical support from the Lamini team and a community forum...}}	 & \textit{\textcolor{red}{For troubleshooting and optimization of Lamini AI models, developers can refer to the Lamini AI documentation, which provides guidelines and best practices for model training and deployment. }\textcolor{darkgreen}{ 
 Additionally, the Lamini community forum and support channels offer a platform for developers to ask questions, share knowledge, and seek assistance from other users and Lamini experts...}} \\
\hline
\textit{How does Lamini's ``add\_data()" function complement the ``submit\_job()" function in enhancing the model's performance and learning capabilities?}	& \textit{\textcolor{darkgreen}{... Together, they work in tandem to ensure that the model not only receives diverse and ample data for training but also applies this information to learn and improve its accuracy and efficiency in handling different tasks.}}	 & \textit{\textcolor{darkgreen}{... function by providing the model with additional training data to enhance its performance and learning capabilities...}}
\\ 
\hline
\end{tabular}
\label{tab:mixed_question}
\end{table*}

Using the manually selected subset of questions-answer pairs from the test dataset, we create 38 MHQA stype question-answer pairs using the method discussed in Section \ref{sec:recall}. Prompting the fine-tuned model with $Q_3$, we compare the generated response with the gold response $A_3$. \\
\\
We fine that most of the trends observed in Section \ref{sec:r-mem} (memorization) holds true for recall as well. The generated responses are almost 1.5 times as long as the gold responses (67.13 words vs 44.86 words per response), and the generated response uses more domain specific words than the gold response (2.13 vs 1.32). Even the correctness of the response follows the trend, where 23.68\% of the responses are wrong, 28.9\% of the responses are correct with extra details, and the remaining 47.36\% of the responses are correct.





Table \ref{tab:mixed_question} presents examples of model generated responses for the recall task. The first row illustrates a scenario where the model gets the answer wrong. The second row shows the model generating the correct answer, but including extra details to its response. Finally, the third row demonstrates a case where the model gets the answer correct (without including extra details).




Hence, we find that the fine-tuned model can satisfactorily answer complex MHQA style questions, while generating longer responses and using important domain specific words more frequently than expected. Note that, although we create complex questions by combining test set questions, the test set questions used for creating these questions were very similar to the train set questions. 

\subsection{Reasoning}
While Sections \ref{sec:r-mem} and \ref{sec:r-recall} has shown that the fine-tuned model can answer questions which are similar to the train set questions, in this section we analyze whether the model actually learns the newly introduced entities or hallucinates. 

\subsubsection{Filtering Newly Introduced Entities}

For performing the three stage filtering process proposed in \ref{sec:new}, we consider each QA pair as a document. Therefore, we have a total of 1260 documents from which we select the top-1000 tf-idf words. The tf-idf scores for these top-1000 words ranges from 1.166 to 103.728, with the word ``lamini" achieving the highest score. While the tf-idf filtering include words such as \{lamini, edit\_config, debiasing, pytorch\}, it also includes common words such as \{text, model, help, engine\}. \\
\\
The second stage of filtering helps remove the common words, and only 168 out of the 1000 tf-idf words are selected. Many of the filtered entities are specific to Lamini such as \{submit\_job, edit\_config, test\_cache\} but contains other domain specific words such as \{gpt4, huggingface, pytorch\}. The third stage helps remove all the domain specific words and only select Lamini functions. Only 39 out of the 168 entities are selected in this stage. Entities such as \{laminiai, diamos, linkerd\} which are newly introduced and specific to Lamini but not function calls are also removed. 

Further, we split the resultant entities into two categories based on whether their function can be deduced from their name:
\begin{itemize}
    \item Functions whose descriptions can be deduced from their are classified under the ``common entities" class. Ex: \{add\_data, cancel\_job, model\_name\}
    \item And all the other entities are classified under ``specific entities" class. Ex: \{full\_balance\_dataset, test\_cache, improve\}
\end{itemize}

We present the results for both the common and specific entities separately, in addition to presenting all entities together. Out of the 39 entities, 23 are common and 16 are specific. 

\subsubsection{Evaluating Reasoning}

We perform the four stage evaluation process proposed in section \ref{sec:eval_hal} on the 39 filtered entities. The results of the analysis is presented in Table \ref{tab:eval_hal}. \\
\\
\textbf{A. Understanding:} While 86.95\% of common entities pass the understanding test, only 18.75\% of the specific entities pass the same test. This shows that the fine-tuned model cannot learn new entities with uncommon function names. \\
\\
\textbf{B. Retrieval:} Interestingly, out of the 3 specific entities that pass the understanding test, none of them pass the retrieval test. Therefore, the model has poor understanding or retrieval capabilities on newly introduced entities if their function names are not common. However, for the common entities the model achieves a retrieval rate of 45\%. \\
\\
\textbf{C. Stress Test:} The stress test helps in evaluating whether the model actually learns the common entities or only uses their names to deduce their descriptions and answer questions. We find that out of the 9 common entities that pass the retrieval test, only one of then passes the stress test. As the model gets easily confused and chooses the wrong function calls, we conclude that even with common entities the model does not actually learn the functions.\\
\\
\textbf{D. Advanced Reasoning:} As most functions fail the previous tests, we are left with only one function to perform the reasoning test. The results of the reasoning tests are: \\

\textit{Q1: Are both .query() in H2O.ai and filter\_fn in Lamini used for filtering data? \\
A1: \textcolor{darkgreen}{Yes, both H2O.ai's .query() and Lamini's filter\_fn are used for filtering data. .query() in H2O.ai is used to create a query that filters the data, while filter\_fn in Lamini is a function that takes in a data generator and returns a filtered data generator. Both are used to select specific data points that meet certain criteria, such as filtering ...} \\
\\
Q2: Are there any similarities between .query() in H2O.ai and filter\_fn in Lamini? \\
A2: \textcolor{red}{There is no similarity between the .query() in H2O.ai and filter\_fn in Lamini as they serve different purposes ...}}

We can observe that while the model has recall for both the entities as it correctly recalls the function descriptions for both .query() and filter\_fn. However, the model is not able to reason the similarities between the two functions. Hence, the model did not demonstrate reasoning ability in this evaluation. None of the entities, common or specific, pass all four stages of testing. 

\begin{table*}[t]
\centering
\caption{Evaluating Hallucinations in Newly Introduced Entities}
\begin{tabular}{|c|c|c|c|c|c|}
\hline
\textbf{Entity Type} & \textbf{Total}            & \textbf{\begin{tabular}[c]{@{}c@{}}A. \\ Understanding\end{tabular}} & \textbf{\begin{tabular}[c]{@{}c@{}}B.\\ Retrieval\end{tabular}} & \textbf{\begin{tabular}[c]{@{}c@{}}C.\\ Stress Test\end{tabular}} & \textbf{\begin{tabular}[c]{@{}c@{}}D. \\ Advanced Reasoning\end{tabular}} \\ \hline
Common               & 23                        & 20                                                                   & 9                                                               & 1                                                                 & 0                                                                \\ \hline
Specific             & 16                        & 3                                                                    & 0                                                               & 0                                                                 & 0                                                                \\ \hline
All                  & {\color[HTML]{343434} 39} & 23                                                                   & 9                                                               & 1                                                                 & 0                                                                \\ \hline
\end{tabular}
\label{tab:eval_hal}
\end{table*}

\section{Conclusion}

This study evaluates whether domain-specific fine-tuned language models would be robust against hallucination. Our findings show that LLMs are adept at answering questions similar to the training data, achieving accuracies as high as 80\%. The model also demonstrates a tendency to include extra information (learned during pre-training) in its answers in around 40\% of its responses. However, the model's ability to accurately reason about and recall new domain-specific information remains limited, leading to instances of hallucination. On newly introduced entities with uncommon names, the model achieved 0\% retrieval in our evaluation. In addition, the model achieved 0\% performance on our advanced reasoning evaluation across both common and specific entities. Taken together, these results suggest that, in our experimental setting, fine-tuning alone was insufficient for robust recall and reasoning over newly introduced domain-specific information. These findings motivate caution when using fine-tuning by itself as a mechanism for injecting new knowledge into a language model. 






Our experiment results suggest caution in relying on fine-tuning alone for introducing new information to a language model, as it fails to adequately recall and reason about the new information. There is still further exploration to be done to address the challenge of hallucination.

\subsection{Limitations}
The main limitation of our study is the small size of the Lamini dataset. Because current LLMs train on copious amounts of data, it was challenging to find data that would be unseen to the chosen LLMs. The Lamini dataset only had 1.26k question-answer pairs, which may not be substantial for the model to actually learn the new information. We only evaluated the smallest version of the LLama model with 7 billion parameters. This was purposefully chosen for the pragmatic simulation of an organization's domain-LLM. An organization likely has a dataset similar to the LaminiDocs and would desire a manageable, practical LLM for its purposes, where larger size models may be more cumbersome than useful. More observations about the domain-specific entities with a larger model may yield better results.

\subsection{Future Work}
While fine-tuning may cause LLMs to still be susceptible to hallucinate on new information, early and current research finds that incorporating context into the prompt of the LLM improves performance of maintaining factual correctness \citep{brown2020fewshot}. Retrieval Augmented Generation (RAG) \citep{Lewis2021} has been proposed as another solution to curtail hallucinations. Our methodology incorporated RAG-based techniques but we did not conduct comprehensive testing. Further exploration and comparison to fine-tuning techniques is needed to propose a best practice to prevent hallucinations.

\subsubsection*{Author Contributions}
The authors have contributed equally to the study.


\label{refs}
\bibliography{references}
\bibliographystyle{iclr2021_conference}


\end{document}